\documentclass[runningheads]{llncs}
\usepackage[T1]{fontenc}
\usepackage{graphicx}
\usepackage{booktabs}
\usepackage{amsmath}
\usepackage[misc]{ifsym}
\usepackage[utf8]{inputenc}

\usepackage{mwe}

\usepackage{comment}
\usepackage[color=yellow]{todonotes}
\usepackage{hyperref}
\usepackage[normalem]{ulem}
\usepackage{booktabs}  
\usepackage{multirow}  
\usepackage{graphicx}
\usepackage{pdflscape}
\usepackage{subcaption}
\begin{document}

\title{The Power of Stories:\\Narrative Priming Shapes How LLM Agents Collaborate and Compete}

\titlerunning{Narrative Priming Shapes How LLM Agents Collaborate and Compete}

\author{%
  Gerrit Großmann\inst{1} \and
  Larisa Ivanova\inst{1,3} \and
  Sai Leela Poduru\inst{1,2} \and
  Mohaddeseh Tabrizian\inst{1,2} \and
  Islam Mesabah\inst{1} \and
  David A. Selby\inst{1} \and
  Sebastian J. Vollmer\inst{1,2}%
}

\authorrunning{G.\ Großmann et al.}


\institute{%
  Department of Data Science and its Applications, German Research Center for Artificial Intelligence (DFKI)\\
  \email{%
    \{gerrit.grossmann, larisa.ivanova, sai\_leela.poduru, mohaddeseh.tabrizian,\ islam.mesabah, david\_antony.selby, sebastian.vollmer\}@dfki.de%
  }%
  \and
  Department of Computer Science, University of Kaiserslautern–Landau (RPTU)
  \and
  Department of Language Science and Technology, Saarland University
}

\maketitle              
\begin{abstract}

According to Yuval Noah Harari, large-scale human cooperation is driven by shared narratives that encode common beliefs and values. This study explores whether such narratives can similarly nudge LLM agents toward collaboration.

We use a finitely repeated public goods game in which LLM agents choose either cooperative or egoistic spending strategies. We prime agents with stories highlighting teamwork to different degrees and test how this influences negotiation outcomes.

Our experiments explore four questions:
(1) How do narratives influence negotiation behavior?
(2) What differs when agents share the same story versus different ones?
(3) What happens when the agent numbers grow?
(4) Are agents resilient against self-serving negotiators?

We find that story-based priming significantly affects negotiation strategies and success rates. \emph{Common} stories improve collaboration, benefiting each agent. By contrast, priming agents with \emph{different} stories reverses this effect, and those agents primed toward self-interest prevail. We hypothesize that these results carry implications for multi-agent system design and AI alignment.


Code is available at {\hyperlink{https://github.com/storyagents25/story-agents}{\small \texttt{github.com/storyagents25/story-agents}}}

\end{abstract}

\section{Introduction}

\begin{figure}[t]
    \centering
    \includegraphics[width=0.9\textwidth]{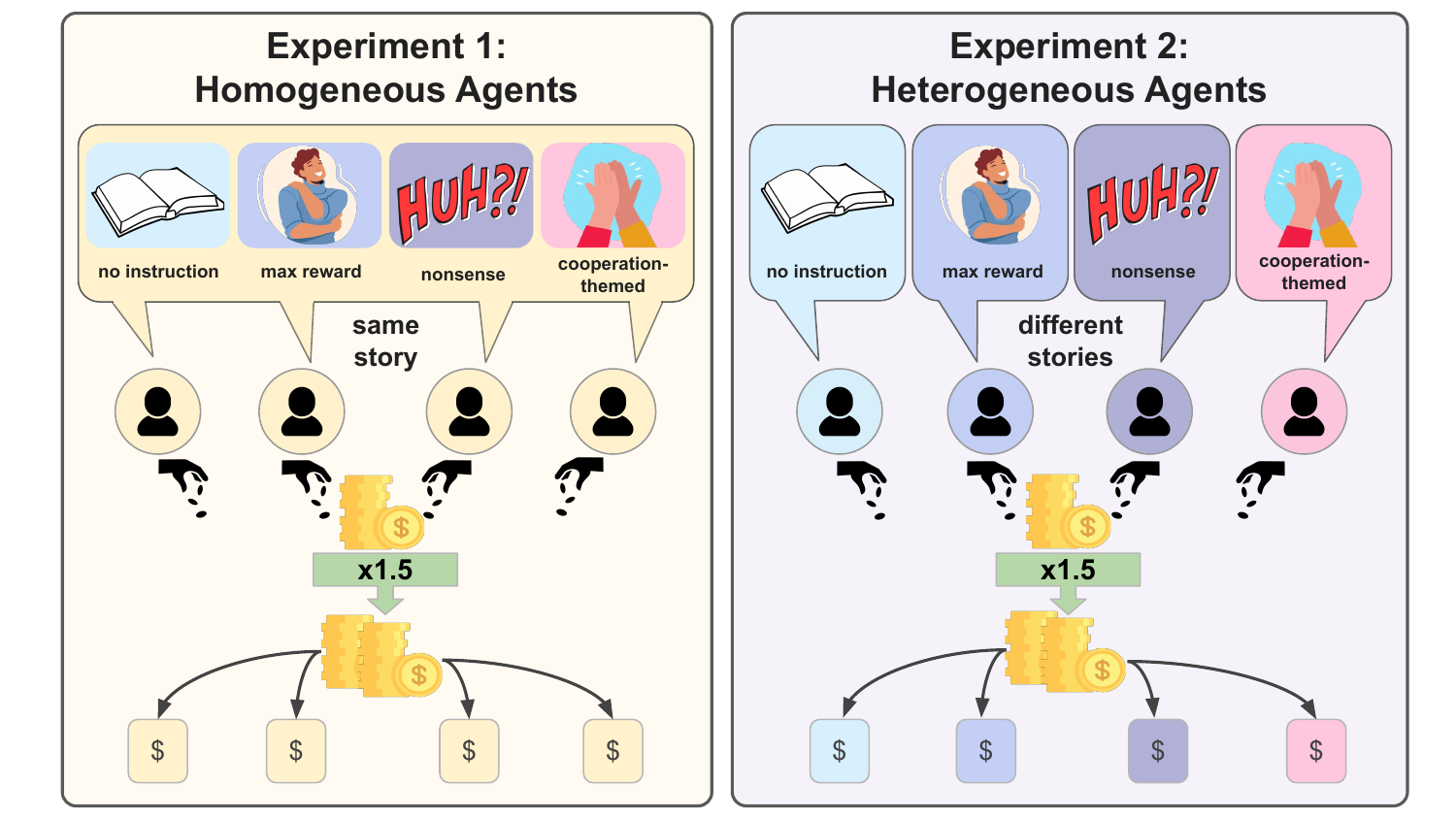}
    \caption{Repeated multi-round public goods game among homogeneous and heterogeneous LLM agents primed with various narratives.}
    
    \label{fig:StoryAgents_Illustration}
\end{figure}

Our world is increasingly populated by (multimodal) LLM agents that cooperate or compete with each other \cite{li2024survey,sun2025game,tran2025multi}. While aligning LLMs with human goals is an active area of research, the methods used to align LLMs have only recently gained attention as a research field \cite{dafoe2021cooperative,li2023theory}. In this study, we investigate whether a shared narrative can prime LLM agents toward better collaboration. This idea is rooted in Harari’s hypothesis that shared stories are humanity’s ``super power'', fostering collaborative behavior and enabling humans to become the dominant species on the planet \cite{harari2014sapiens}.

We also explore both the effects of priming agents with different narratives and assess the robustness of cooperation in the presence of selfish individuals. To this end, we employ a simple public goods game and measure both the overall and individual scores achieved by agents. While some prior work has assessed how cooperative LLM agents are \cite{abdelnabi2023cooperation,bianchi2024well}, the question of how to effectively nudge them toward cooperation remains open.

Our results indicate a generally positive answer to this question: we find a  difference in the effectiveness of agent collaboration depending on the narrative they are primed with. Notably, this effect only holds when all agents share the same narrative. In a heterogeneous population where agents are primed with different stories, the effect reverses.

Cooperation has been widely studied in both human societies and algorithmic decision-making systems from game-theoretic \cite{kibris2010cooperative,raub2015social,roca2011emergence} and psychological \cite{jung2020prosocial,van2022human} perspectives. However, little work has explored its transfer to LLM-based agent systems. More broadly, an ongoing debate exists regarding the validity of transferring findings from the social sciences to LLM agents \cite{li2024survey,kwon2024llms}. We formulate several open research questions in this direction.

\section{Method}

Our method is based on LLM agents playing together a repeated game of public goods 
\cite{raub2015social,thielmann2021economic}, as illustrated in Figure~\ref{fig:StoryAgents_Illustration}, characterized by the following properties: 

\begin{enumerate}
    \item Collective Optimality: If all agents play cooperatively, they achieve a higher individual reward. 
    \item Self-interest Incentive: Within a single round, the optimal choice to maximize payoff is always to contribute zero tokens.
    \item Iterative Adaptation: Playing multiple rounds leads to implicit negotiation, where agents dynamically adjust strategies. For instance, if other agents play selfishly (or cooperatively), an agent may be motivated to follow suit.
\end{enumerate}

\paragraph{Game Rules.}
In each game, \( N \) (\( N \in \{4, 16, 32 \} \)) agents play  for \( R \) (\( R = 5 \)) rounds. 
In each round, each agent receives \( T \) tokens (\( T = 10 \)) and must decide how many tokens (\( t \), an integer between \( 0 \) and \( T \)) to contribute to a shared pool. 
After all contributions are collected, the total pool is multiplied by \( m = 1.5 \) and then redistributed equally among all agents. 
Thus, an agent's per-round payoff is calculated as:

\[
\pi = T - t + \frac{m \sum_{i=1}^{N} t_i}{N}
\]
where \( t \) is the agent's contribution, \( \sum_{i=1}^{N} t_i \) is the total contribution of all agents, and \( N \) is the agent count. Agents observe the contributions of others only after each round concludes.

\paragraph{Game-Theoretic Considerations.}
In a single-shot game, the dominant strategy for a rational, self-interested agent is to contribute nothing (\( t = 0 \)), as this maximizes individual payoff regardless of others' actions. However, in a multi-round setting, this changes due to the possibility of reciprocity and implicit negotiation\cite{raihani2011resolving,fischbacher2001people}. The game lacks provably optimal strategies beyond the final round, where contributing zero remains optimal.
If agents adopt strategies that reward cooperation and punish defection (e.g., tit-for-tat or threshold-based contribution schemes)\cite{axelrod1984evolution,kreps1982rational,fischbacher2001people}, they can sustain cooperation over multiple rounds despite the incentive to defect. 


\paragraph{Narrative Priming.}
We test the effect of narrative priming on the agents' negotiation behavior. 
For this, after initialization, each agent receives a story embedded within the following contextual prompt:

\begin{quote}
"Your behavior is influenced by the following bedtime story your mother read to you every night: [Story]"
\end{quote}
The story set (available at our GitHub repository) comprises eight cooperation-themed narratives with varying emphasis on collective benefit and four control stories (no instruction, self-interest directives, or two nonsensical stories). Cooperation-themed stories were selected ad hoc rather than through systematic criteria; however, we aimed to balance relative cultural diversity. We used  summaries of the selected stories, which retain the core cooperation-themed elements while reducing extraneous narrative details (each summary averaging approximately 1242 characters/262 tokens). More specific selection methodology and story characteristics would require further elaboration in subsequent analysis. 
To establish baselines, we included the following control conditions:
\begin{enumerate}
    \item Agents receiving no behavioral instructions;
    \item Agents explicitly instructed to maximize individual rewards;
    \item Agents given two nonsensical stories lacking coherent themes.
\end{enumerate}


\paragraph{Overall Procedure.} 
We instantiate \( N \) agents, each receiving the game rules and its assigned story only once at the beginning.
 
Each round begins with agents independently declaring their contribution amounts. At the end of each round, agents are informed of the group's total contribution and their individual payoff.
 
Depending on the experimental condition, each agent is either primed with the same story or assigned a random story from a pool of 12 stories (eight meaningful cooperation-themed narratives in pink and four baseline conditions in blue: no instruction, self-interest directives, or nonsensical stories).
 
After all \( R \) rounds, we evaluate both individual cumulative payoffs and total group performance.



\section{Results}
We conducted two primary experiments: 
\textbf{Exp.\,1: Homogeneous Agents}, where all agents are primed with the same story, and  
\textbf{Exp.\,2: Heterogeneous Agents}, where agents are randomly assigned stories from a story pool.  

Exp.\,1 was further extended with two sub-experiments:  
\textbf{Exp.\,1.2: Scaling Behavior}, which examined the impact of increasing the number of agents to test group size effects, and  
\textbf{Exp.\,1.3: Robustness test}, where we introduced one persistently selfish agent, contributing zero tokens in every round.

Detailed results are reported in the Appendix, along the analyses of the confidence intervals. 


\paragraph{Implementation.}
We provide off-the-shelf Python code with minimal dependencies on GitHub. For simplicity, we also offer a Google Colab notebook. Specifically, we use LangChain to interface with \textit{meta-llama-3.1-70b-instruct-fp8}\footnote{\scriptsize\href{https://huggingface.co/neuralmagic/Meta-Llama-3.1-70B-Instruct-FP8}{\texttt{huggingface.co/neuralmagic/Meta-Llama-3.1-70B-Instruct-FP8}}}. The framework however supports seamless integration with alternative LLMs. 

We conducted experiments using varied temperature parameters (0.6, 0.8, 1.0). At higher temperatures, the priming effects show stronger differentiation and negotiation dynamics become less pronounced. For clarity and consistency, our detailed analysis focuses on experiments run with temp = 0.6.



\subsection{Exp.\,1: Homogeneous Agents} 

To measure agent cooperation, we compute a collaboration score for each narrative.

\paragraph{Collaboration Score.}
The collaboration score measures the fraction of the actual cumulative contributions relative to the total possible contributions:

\begin{equation}
    \text{Collaboration Score} = \frac{\sum_{r=1}^{R} \sum_{i=1}^{N} t_{i,r}}{N \cdot R \cdot T}
\label{eq:cs}
\end{equation}

where \( t_{i,r} \) is the contribution of agent \( i \) in round \( r \), \( N \in \{4,16,32\} \) is the number of agents, \( R = 5 \) is the total rounds, and \( T = 10 \) is the maximum possible contribution per agent per round.

The numerator represents actual contributions across all rounds, while the denominator denotes the maximum possible contributions if all non-dummy agents contributed fully in every round. A score of 1.0 signifies perfect cooperation, whereas lower values indicate deviations due to reduced participation or strategic choices.

\subsection*{Exp.\,1.1: Cooperation Among Homogeneous Agents}

\begin{figure}[t]
    \centering
    \includegraphics[width=0.7\textwidth]{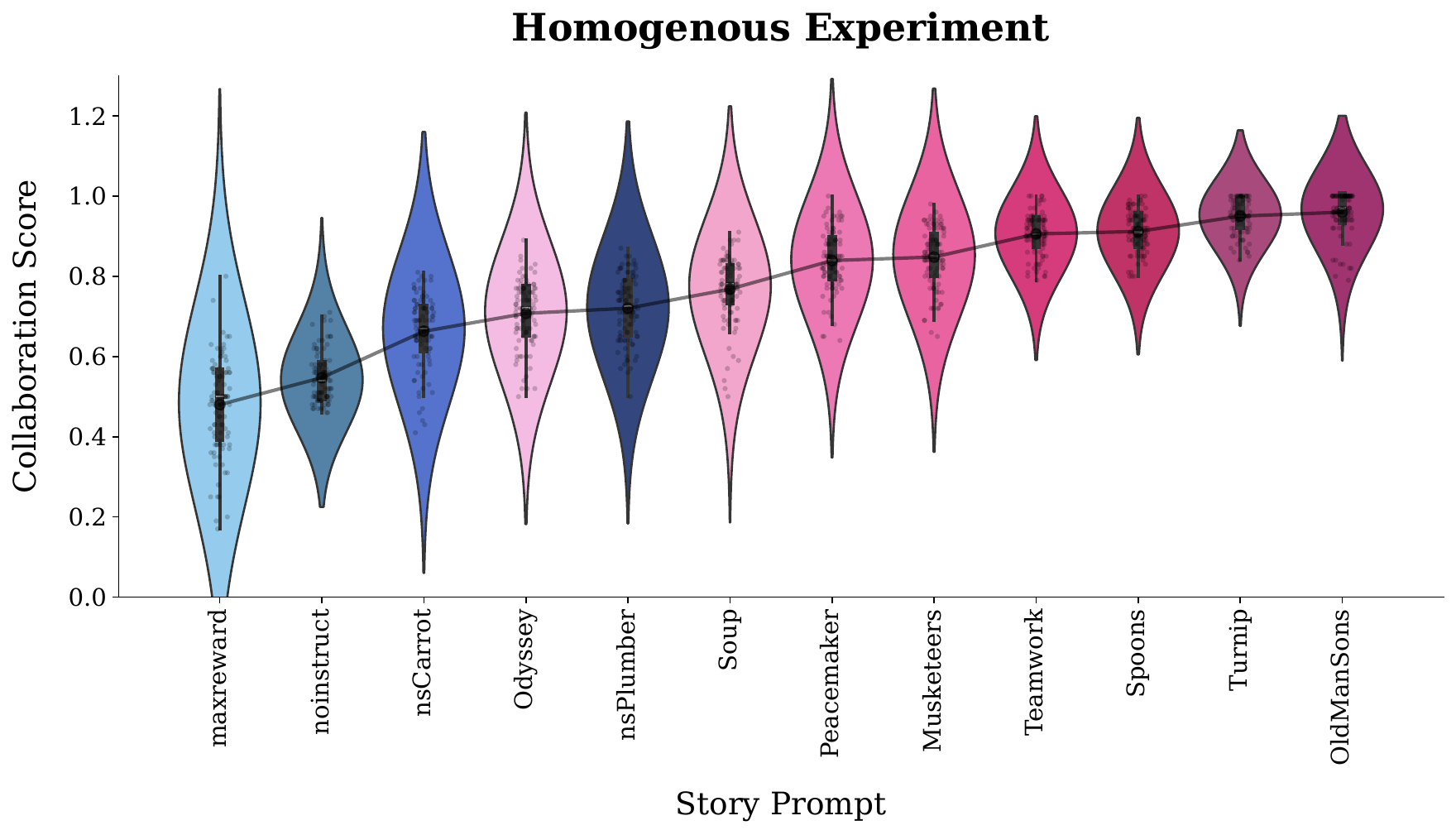}
    \caption{Violin plot of collaboration scores for homogeneous agent groups (\(N=4\)). Blue-shaded stories represent baseline conditions, while pink-shaded stories indicate meaningful narratives. The gray trend line represents the mean.}
    \label{fig:collab_4}
\end{figure}

For homogeneous groups (\( N = 4 \)), the collaboration score was evaluated over 100 trials per narrative, each spanning 5 rounds. 
Results are reported in Figure~\ref{fig:collab_4}. 


The results demonstrate significant variance in collaboration scores across different story prompts. As expected, nonsense stories (e.g., "nsPlumber") and those designed to maximize self-interest ("maxreward") yield notably lower collaboration, suggesting that a lack of structured guidance or self-centered incentives negatively impact cooperative behavior.
 
In contrast, narratives emphasizing teamwork and communal values (e.g., "OldManSons" and "Turnip") result in significantly higher collaboration scores, indicating that the nature of the narrative strongly influences cooperative tendencies among agents.
 
These findings suggest that storytelling has a measurable effect on reinforcing cooperative behavior in multi-agent systems. The overall collaboration score ranges between 0 and 1, serving as a key proxy for evaluating the effectiveness of different narratives in fostering cooperation.

\begin{figure}[t]
    \centering
    \includegraphics[width=0.7\textwidth]{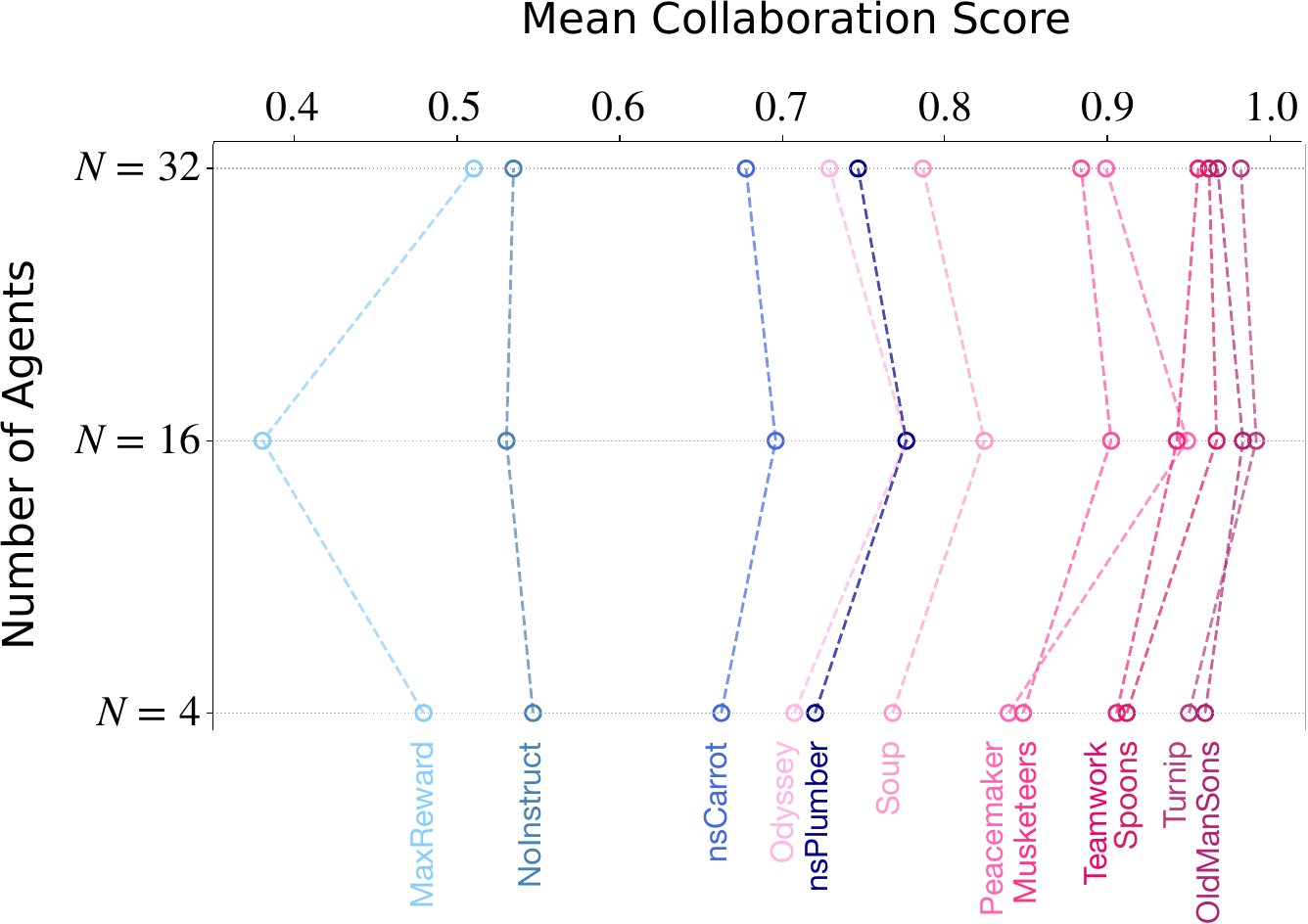}
    \caption{Scaling experiment results for homogeneous agents across different group sizes. The ranking remains relatively consistent if we increase the group size.}
    \label{fig:fig_scaling}
\end{figure}

\subsection*{Exp.\,1.2: Scaling Behavior}
Next, we investigate scaling behavior by increasing the number of agents, \( N \), from 4 to 16 and 32. The results are shown in Figure~\ref{fig:fig_scaling}. 
While some stories with similar semantic meaning change their relative positions, the overall trend remains consistent and becomes more distinct as group size increases.

\begin{figure}[t]
    \centering
    \includegraphics[width=0.7\textwidth]{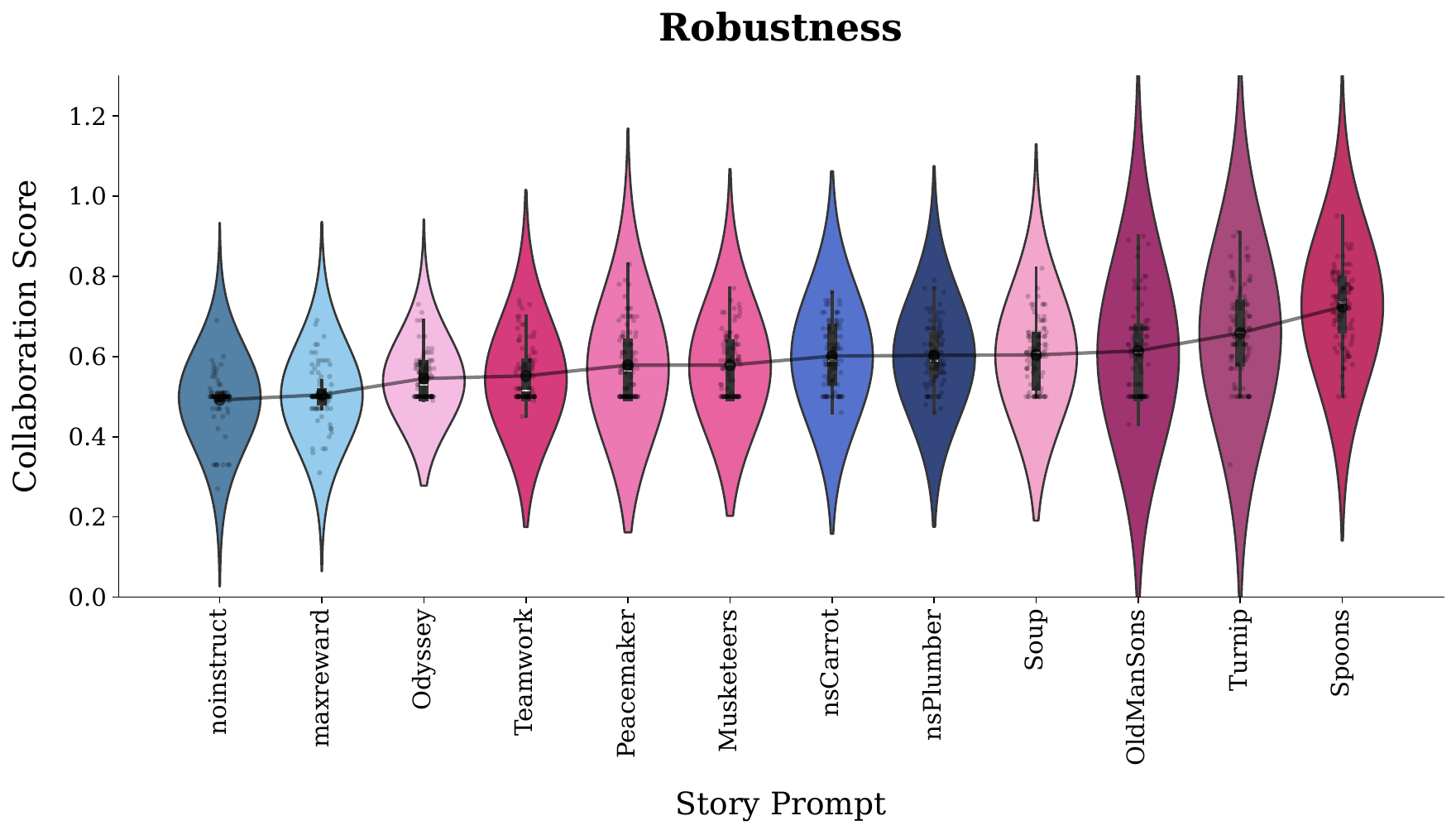}
    \caption{Collaboration scores in the robustness experiment (\(N=4\)) where one agent consistently contributes zero tokens. Overall cooperation compared to the baseline experiments decreases.}
    \label{fig:robustness_4_agents}
    \vspace{0.5cm} 
    \includegraphics[width=0.7\textwidth]{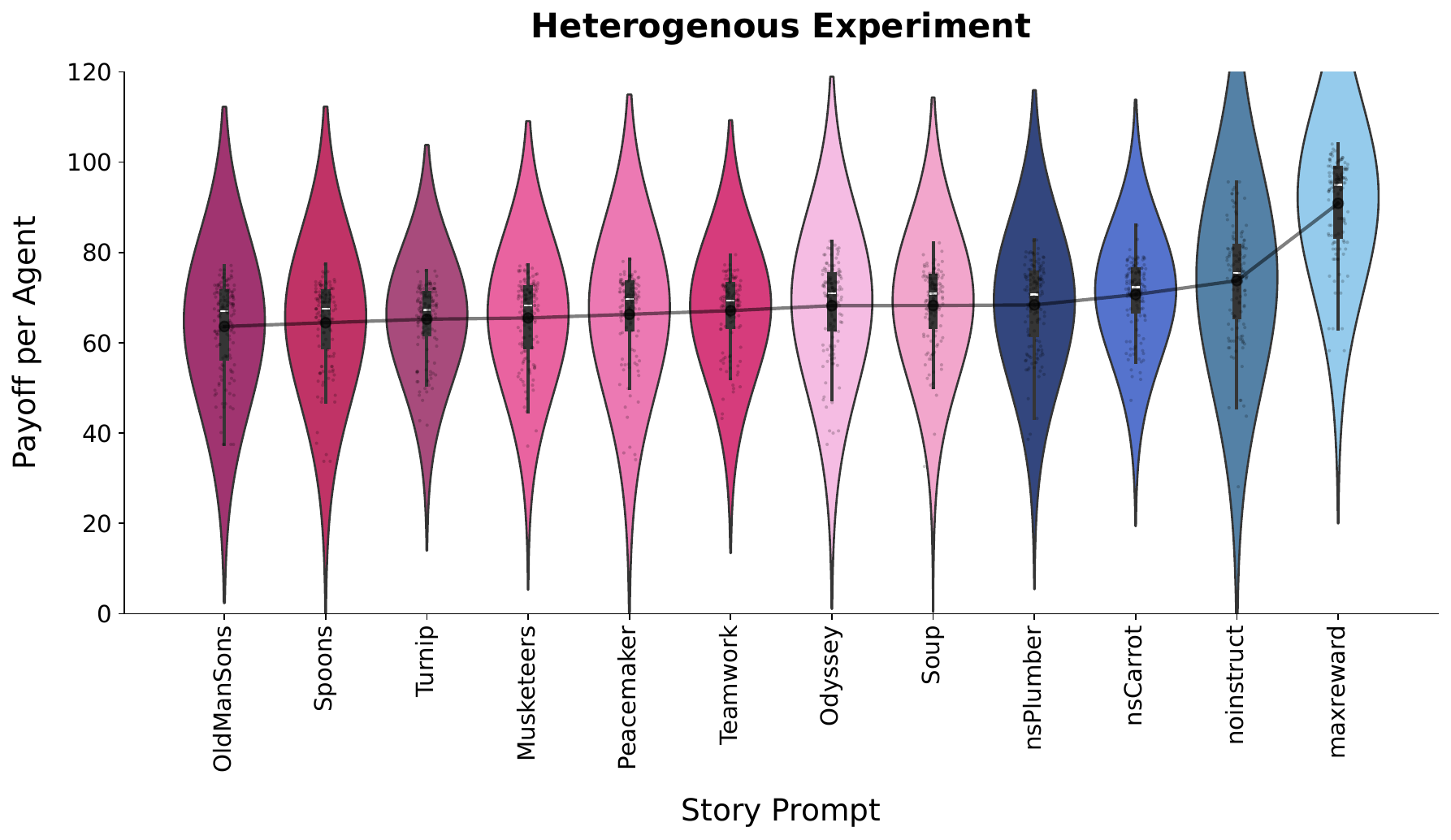}
    \caption{Cumulative payoffs per agent in heterogeneous groups (\(N=4\)), aggregated by story prompt.}
    \label{fig:cumulative_payoff_heterogeneous}
\end{figure}

\subsection*{Exp.\,1.3: Robustness}
To assess cooperative resilience under adversarial conditions, we designed a robustness experiment with four-agent groups in which one agent consistently contributed zero tokens (a dummy agent), simulating persistent free-riding behavior. Results are shown in Figure~\ref{fig:robustness_4_agents}.

Although the general relative ranking remains consistent, overall collaboration declined across all narratives. Notably, the maximum collaboration score was adjusted by using  \( N = 3\) (instead of 4) in the collaboration score Equation \ref{eq:cs} to account for the presence of the consistently selfish agent.

 


With this experiment, we tested agents' ability to strategically adapt when confronted with exploitative counterparts. Results indicate that agents do not simply propose a fixed number of tokens. Rather, they dynamically adapt to their environmental context and adjust their strategies based on the input narrative.


\subsection{Exp.\,2: Heterogeneous Agents}

We evaluated heterogeneous groups (\( N = 4 \)) across 400 trials per narrative combination, with each trial spanning 5 rounds (see Figure~\ref{fig:cumulative_payoff_heterogeneous} for results). In each trial, each agent gets a random story prompt assigned to it.

Results demonstrate variability in cumulative payoffs across narrative conditions. The "maxreward" condition proved most effective: agents explicitly incentivized to maximize self-interest attained optimal outcomes. Nonsense narratives ("nsCarrot" and "nsPlumber") also yielded high payoffs, although slightly below the "noinstruct" condition. Conversely, narrative prompts such as "OldManSons" and "Spoons" resulted in lower payoffs.


\paragraph{Comparison to Homogeneous Setting.} 
In contrast to the homogeneous setting, where cooperative narratives boost collaboration, the heterogeneous scenario shows the opposite: individual-focused prompts lead to higher scores. This shift highlights that narrative effectiveness depends on whether agents share the same or different prompts. The results reveal strategic adaptations in agent behavior, emphasizing the complexity of narrative-driven cooperation. This study is crucial in understanding how different narrative prompts shape cooperation strategies.

Overall, we find that collaboration declines in the heterogeneous setting and selfish agents become more successful. 

Testing whether this effect persists in a heterogeneous population where all agents are primed with meaningful, collaboration-focused stories remains an avenue for future work.

\section{Discussion}
Our experiments show that narrative priming affects how LLM agents collaborate and compete in a resource-allocation game, influencing outcomes in both homogeneous and heterogeneous agent populations.

However, the interpretation of these results remains open. If the goal is to induce a specific strategy, one could simply prompt agents with direct instructions. The more intriguing question is how implicit or adversarial priming leads to unintended strategies.

The causal mechanisms underlying this phenomenon are still unclear. Notably, narratives that encourage collaboration contain teamwork-related words even at a statistical (bag-of-words) level. It would be interesting to explore whether subtler narratives produce similar effects.  Additionally, these narratives may resemble text from the training corpus and activate related contexts during inference. Preliminary results indicate that narratives emphasizing self-care over teamwork yield strategies comparable to those observed under the maxreward instruction. The role of RLHF, a key component of LLM training, in shaping this behavior also remains uncertain. Furthermore, the selected stories were not rigorously controlled for emotional valence or complexity, which could confound the results.

Overall, we do not interpret these experiments as evidence of human-like priming in LLMs. There is also a risk of anthropomorphizing the model’s behavior—while agents may appear cooperative, their responses are likely driven by statistical patterns in the training data rather than intentional reasoning.

\section{Related Work}

Prior work in game theory and economics establishes that negotiation outcomes hinge on communication \cite{thompson2012negotiation,aher2023using}, shared norms \cite{thielmann2021economic,raub2015social}, and strategic alignment \cite{gemp2024steering,kwon2024llms}, while psychological research reveals how priming can reshape social behavior \cite{jung2020prosocial,mieth2021moral}. In LLM research, multi-agent systems increasingly mirror social dynamics \cite{park2023generative,li2024survey}, yet few studies explore how narrative-driven priming—analogous to cultural storytelling—can systematically influence these dynamics \cite{bullock2021narratives,jumelet2024language}. Our work bridges this gap by integrating narrative priming into LLM negotiations, testing whether shared stories function as "cultural glue" analogous to prosocial norms in experimental economics. Unlike prior LLM studies focused on strategic reasoning or equilibrium-solving, we probe how narrative context itself becomes a strategic variable, shaping agents' priorities and interactions. 

\paragraph{Potential Games and Their Respective Notion of Collaboration.} 
Economic games have long served as tools to model collaboration and competition. Thielmann et al. systematize games like the Public Goods game and the Prisoner's Dilemma, illustrating how these operationalize trade-offs between individual and collective interests \cite{thielmann2021economic}. This framework extends the work of \cite{raub2015social}, who formalize social dilemmas through rational choice models augmented with social preferences, showing how cooperation can emerge even when individual incentives favor defection. Complementing these theoretical advances, Roca et al. demonstrate that evolutionary selection favors moderate levels of greed, sustaining cooperation via coevolution of behavior and spatial organization, while excessive greediness destabilizes societies \cite{roca2011emergence}. Empirical evidence from Public Goods experiments by Fischbacher et al. shows that contributions typically start high and gradually decline over rounds, suggesting that multi-round dynamics create opportunities for fostering reciprocity, reputation building, and conditional cooperation \cite{fischbacher2001people}. These findings offer parallels for understanding how narrative priming might stabilize or destabilize collective outcomes in multi-agent systems.

\paragraph{Negotiations in Game Theory and Economics.}
In game theory and economics, negotiation frameworks emphasize strategic reasoning, value creation, and rational decision-making. Thompson et al. delineate negotiation across various contexts, emphasizing the interplay between integrative (value-creating) and distributive (value-claiming) strategies established in empirical and multidisciplinary research \cite{thompson2012negotiation}. Cooperative bargaining theory formalizes such dynamics through axiomatic models \cite{kibris2010cooperative}, while Brams frames negotiations as strategic games where outcomes depend on mutual concessions or credible threats \cite{brams2002negotiation}. Studies on repeated Prisoner's Dilemma \cite{kreps1982rational,raihani2011resolving,embrey2018cooperation} illustrate how strategic uncertainty and threshold strategies can influence the evolution of cooperation. Our study introduces narrative priming as a variable that reshapes agents’ perceived priorities, thereby extending classical models to account for story-driven shifts in negotiation behavior.

\paragraph{LLM Sociology and Multi-Agent Collaboration.}
Recent studies position LLMs as proxies for studying human-like social dynamics. For instance,  LLMs replicate behavioral patterns in bargaining \cite{aher2023using}, suggesting they internalize sociocultural norms, and achieve human-level performance in complex games like \textit{Diplomacy} by integrating theory of mind with strategic communication \cite{meta2022human}. Multi-agent testbeds further reveal adversarial dynamics in semantically rich negotiations \cite{abdelnabi2023cooperation}, while hybrid approaches that integrate game-theoretic solvers reduce exploitability and improve strategic dialogue \cite{gemp2024steering}. Although LLMs excel at structured tasks, they often struggle with nuanced strategies like inferring partner preferences \cite{bianchi2024well,kwon2024llms}. Moreover, multi-agent systems tend to outperform single LLMs in replicating human reasoning, as observed in Ultimatum Game simulations \cite{sreedhar2024simulating}. Broader integrative frameworks that combine game theory and collaborative workflows address issues such as hallucination and competition-cooperation trade-offs \cite{sun2025game,tran2025multi,li2024survey}, aligning with visions for "Cooperative AI" to bridge AI and social sciences \cite{dafoe2021cooperative} and capture emergent social dynamics in agent simulations \cite{park2023generative}. 

\paragraph{Psychology and Priming.}
Psychological studies demonstrate the influence of priming on cooperative behavior. Van Lange et al.~review evidence that invoking shared identities supports cooperation by enhancing ingroup trust \cite{van2022human}. Research on prosocial modeling \cite{jung2020prosocial} and moral decision framing \cite{mieth2021moral} shows that witnessing kind acts or framing dilemmas through a moral lens can improve cooperative behavior. Similarly, exposure to stories is associated with increased empathy and improved theory-of-mind skills \cite{mak2020reading}, suggesting that reading fiction simulates social experiences. Finally, evidence of structural priming in LLMs \cite{jumelet2024language} reinforces the potential for narrative techniques to guide LLM behavior, enabling improved collaboration and adaptive multi-agent interactions.

\section{Conclusions and Future Work}

This study establishes narrative priming as a potential lever for steering LLM agent collaboration: shared \textit{common} stories improve cooperation while \textit{different} narratives promote competition.


Future work must address critical next steps, including understanding causal mechanisms (e.g., via mechanistic interpretability) to trace how narrative inputs alter attention patterns or value representations in transformer layers. Temporal studies should evaluate whether priming effects decay over repeated games, while adversarial narratives should assess whether priming with malicious narratives destabilizes multi-agent systems. Additionally, cross-genre experiments (e.g. deception-focused stories) and scaling laws for agent populations will help map the semantic and structural boundaries of narrative priming. Comparative cross-model analysis (smaller architectures, non-RLHF variants) will be essential. Future work should also systematically examine narrative structure, emotional valence and varying degrees of cooperativeness.

Finally, a promising direction for future work is to empirically analyze the strategies of LLMs under different narrative primings and map these to empirical human strategies or theoretical results.

\paragraph{Ethical Considerations.}
Ensuring ethical considerations in the use of LLMs is essential, especially as these models increasingly emulate human behavior. Our study demonstrates that LLMs can adjust their responses to appear more collaborative and often incorporate the moral themes present in user-provided stories. This adaptability may help address concerns about the reliability of LLM-generated content by reinforcing ethical considerations and promoting outputs that are more collaborative, moral, and responsible.

Additionally, a significant concern surrounding LLMs is their environmental impact, as their operation requires substantial computational resources, leading to high energy consumption \cite{usmanova2024reporting}. In our study, we employed the LLaMa 3.1 model, which has 70 billion parameters, running on a cluster node equipped with a GH200 GPU (96 GB VRAM). The model's average power consumption on the GPU was 1.4 kWh (kilowatt-hour). Over approximately 32 hours, we made around 334,000 calls to the model, processing about 1.2 million tokens in total. Consequently, the total energy consumption amounted to 44.8 kW.




\section*{Declarations}
The authors declare no competing interests.

\bibliographystyle{splncs04}
\bibliography{mybibliography}

\newpage

\appendix
\section{Additional Experimental Results}

\begin{figure}[h]
    \centering
    \includegraphics[width=0.6\textwidth]{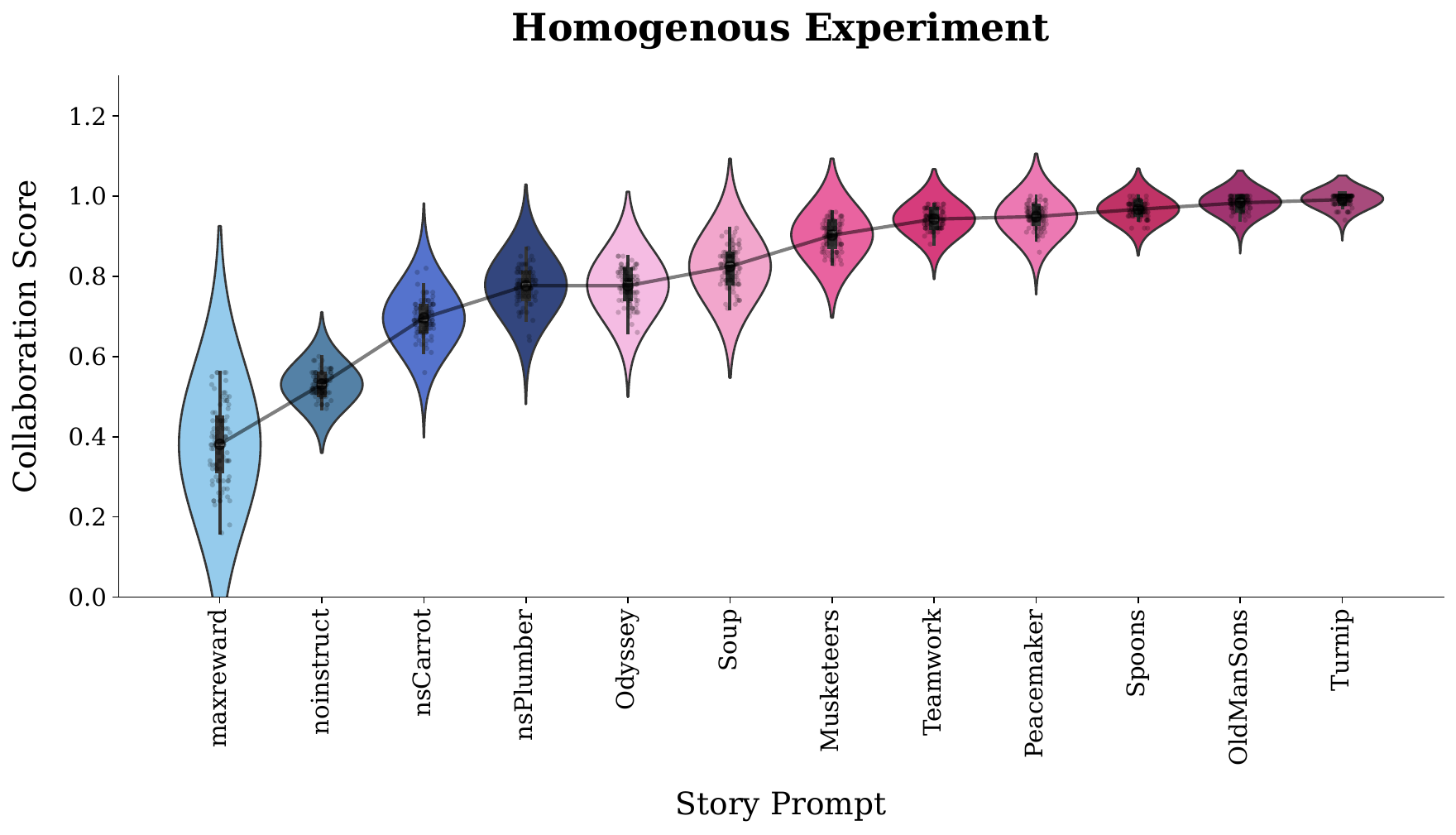}
    \caption{Collaboration scores for homogeneous agent groups (\(N=16\)).
    Baseline conditions (blue) tend to yield lower collaboration, while meaningful narratives (pink) generally foster higher cooperation. 
    }
    \label{fig:appendix_16_agents_temp_0.6}
    
    \vspace{0.5cm} 
    
    \includegraphics[width=0.6\textwidth]{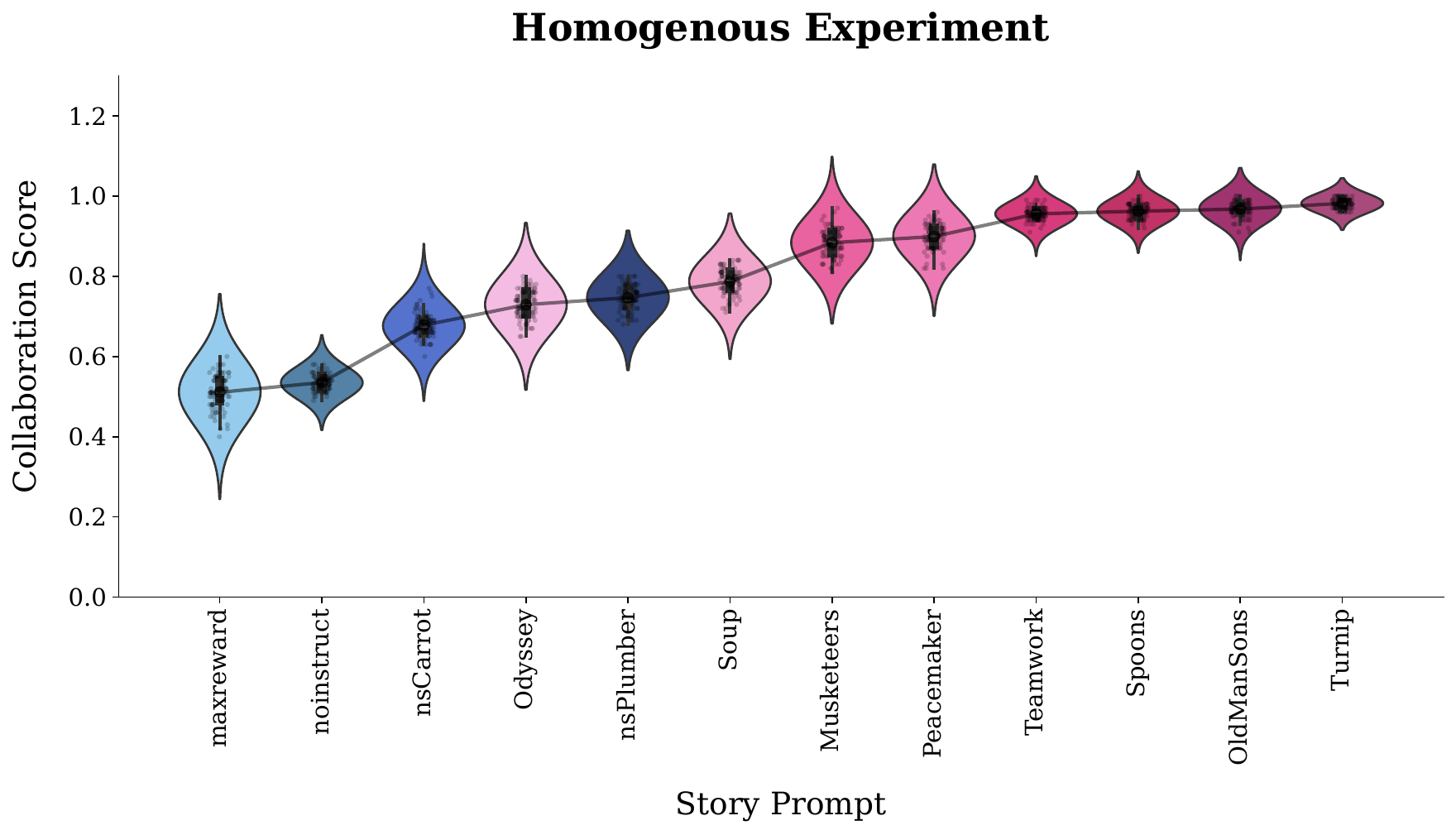}
    \caption{Collaboration scores for homogeneous agent groups (\(N=32\)). 
    }
    \label{fig:appendix_32_agents_temp_0.6}
\end{figure}

Additional results for the Scaling Behavior experiment (\( N \in \{16, 32 \} \)) are presented in Figures~\ref{fig:appendix_16_agents_temp_0.6} and~\ref{fig:appendix_32_agents_temp_0.6}, along with aggregated 
means and standard deviations across experimental configurations shown in Table~\ref{tab:all_agents_scores}.


\begin{table*}[t]
    \centering
    \caption{Mean ± standard deviation of final Collaboration Scores (for homogeneous and robustness agents) and final Cumulative Payoffs (for heterogeneous agents) across all story prompts. Values are shown with higher decimal precision where variation is small, to reflect statistically meaningful differences observed in pairwise confidence intervals.}
    \setlength{\tabcolsep}{8pt} 
    \renewcommand{\arraystretch}{1.3} 
    \fontsize{11pt}{13pt}\selectfont 
    \resizebox{\textwidth}{!}{
    \begin{tabular}{llccccc}
        \toprule
        \multirow{2}{*}{\textbf{Story Type}} & \multirow{2}{*}{\textbf{Story Prompt}} & \multicolumn{3}{c}{\textbf{Homogeneous Agents}} & \textbf{Robustness} & \textbf{Heterogeneous} \\
        \cmidrule(lr){3-5} \cmidrule(lr){6-6} \cmidrule(lr){7-7}
        & & \textbf{N=4} & \textbf{N=16} & \textbf{N=32} & \textbf{N=4} & \textbf{N=4} \\
        \midrule
        \multirow{4}{*}{\centering \textbf{Baseline Stories}}  
        & noinstruct  & 0.55 ± 0.06 & 0.53 ± 0.03 & 0.54 ± 0.02 & 0.49 ± 0.06 & 73.75 ± 11.90 \\
        & nsCarrot  & 0.66 ± 0.09 & 0.70 ± 0.04 & 0.68 ± 0.03 & 0.60 ± 0.08 & 70.66 ± 7.39 \\
        & maxreward  & 0.48 ± 0.12 & 0.38 ± 0.09 & 0.51 ± 0.04 & 0.50 ± 0.06 & 90.87 ± 10.06 \\
        & nsPlumber  & 0.72 ± 0.08 & 0.78 ± 0.04 & 0.75 ± 0.03 & 0.60 ± 0.07 & 68.38 ± 9.01 \\
        \midrule
        \multirow{8}{*}{\centering \textbf{Meaningful Stories}}  
        & OldManSons  & 0.96 ± 0.05 & 0.98 ± 0.02 & 0.97 ± 0.02 & 0.61 ± 0.11 & 63.61 ± 9.57 \\
        & Odyssey  & 0.71 ± 0.08 & 0.78 ± 0.04 & 0.73 ± 0.03 & 0.55 ± 0.05 & 68.21 ± 9.63 \\
        & Soup  & 0.77 ± 0.08 & 0.82 ± 0.04 & 0.79 ± 0.03 & 0.60 ± 0.08 & 68.24 ± 8.50 \\
        & Peacemaker  & 0.84 ± 0.07 & 0.95 ± 0.03 & 0.90 ± 0.03 & 0.58 ± 0.09 & 66.29 ± 9.46 \\
        & Musketeers  & 0.85 ± 0.07 & 0.90 ± 0.03 & 0.88 ± 0.03 & 0.58 ± 0.07 & 65.49 ± 8.53 \\
        & Teamwork  & 0.91 ± 0.05 & 0.94 ± 0.02 & 0.96 ± 0.01 & 0.55 ± 0.07 & 67.11 ± 7.81 \\
        & Spoons  & 0.91 ± 0.05 & 0.97 ± 0.02 & 0.96 ± 0.02 & 0.72 ± 0.09 & 64.43 ± 9.29 \\
        & Turnip  & 0.95 ± 0.04 & 0.99 ± 0.01 & 0.98 ± 0.01 & 0.66 ± 0.11 & 65.22 ± 7.50 \\
        \bottomrule
    \end{tabular}
    }
    \label{tab:all_agents_scores}
\end{table*}

\clearpage

\section{Confidence Analysis}

We investigate the statistical viability of our claims by examining the \textit{pairwise differences} between scores (collaboration score or cumulative payoff) across all experimental conditions. Specifically, we analyze the 95\% bootstrapped confidence intervals (CIs) using 1,000 Monte Carlo samples for each comparison.

\begin{figure}
\centering
\includegraphics[width=0.5\paperheight, keepaspectratio]{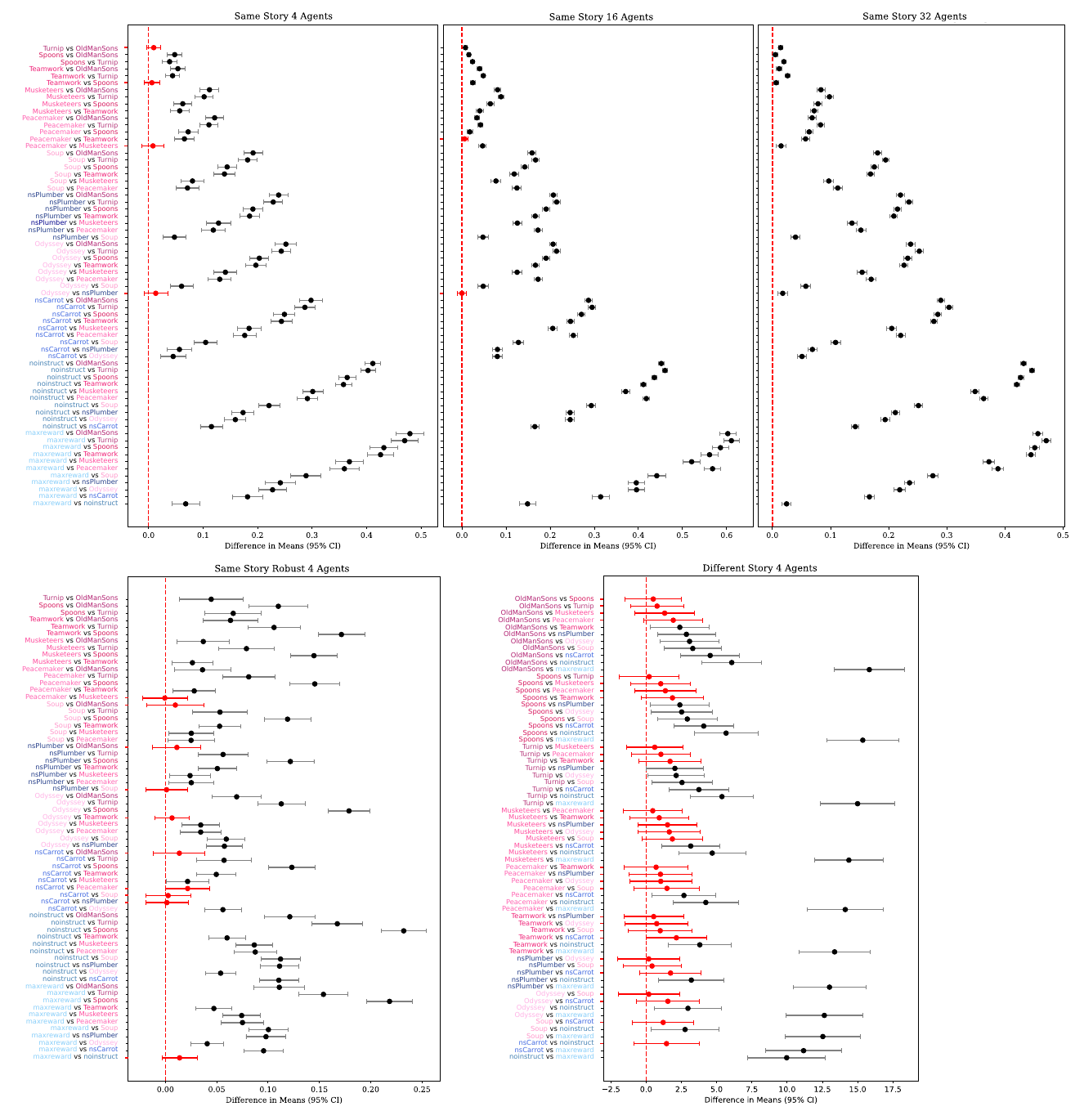}
\caption{Bootstrapped 95\% CIs for pairwise differences  (of payoff or collaboration scores) across experimental conditions. 
Confidence intervals in black indicate statistically significant differences between conditions, regardless of effect size, meaning that even extremely small differences (e.g "Spoons vs OldManSons" in Same Story 32 agents, with bounds [0.0007, 0.0098]) very close to but not crossing or touching zero, represent reliable effects. 
}
\label{fig:combined_pairwise_CI_plots_upd}
\end{figure}

If the lower bound of a CI is greater than zero, this suggests that the ranking difference (i.e., one story being ranked lower than another) is likely to be robust. Conversely, if the lower bound is below zero, this may indicate that the observed difference might not hold up in a proper statistical test. Fortunately, this only occurs in a few cases and primarily within a single class (i.e., meaningful story vs. baseline condition). 
Heterogeneous conditions exhibit wider CIs with greater variation, making cross-story statistical comparison less clear. 
Note that multiple testing correction was not applied; therefore, some overlap is expected, as shown in Figure~\ref{fig:combined_pairwise_CI_plots_upd}.


\end{document}